\documentclass[final]{cvpr}

\usepackage{graphicx}
\usepackage{subcaption}
\newcommand{\our}{GAEL}

\usepackage{amsmath}
\usepackage{amssymb}

\usepackage[pagebackref=true,breaklinks=true,colorlinks,bookmarks=false]{hyperref}

\def\cvprPaperID{7326} 
\def\confYear{CVPR 2021}

\begin{document}

\title{GMM-Based Generative Adversarial Encoder Learning}

\author{%
  Yuri Feigin, Hedva Spitzer and Raja Giryes \\
  School of Electrical Engineering\\
  Tel Aviv University\\
  Israel, 69978 \\
  \texttt{\{yurifeigin@mail, hedva@tauex, raja@tauex\}.tau.ac.il} \\
  }

\maketitle

\begin{abstract}
While GAN is a powerful model for generating images, its 
inability to infer a latent space directly limits its use in applications requiring an 
encoder. Our paper presents a simple architectural setup that combines the generative capabilities of GAN with an encoder. 
We accomplish this by combining the encoder with the discriminator using shared weights, then training them simultaneously using a new loss term.
We model the output of the encoder latent space via a GMM, which leads to both good clustering using this latent space and improved image generation by the GAN. Our framework is generic and can be easily plugged into any GAN strategy. In particular, we demonstrate it both with Vanilla GAN and Wasserstein GAN, where in both it leads to an improvement in the generated images in terms of both the IS and FID scores. 
Moreover, we show that our encoder learns a meaningful representation as its clustering results are competitive with the current GAN-based state-of-the-art in clustering.\footnote{Source code will be released upon publication.}
\end{abstract}
\section{Introduction}

\begin{figure*}
\centering
\includegraphics[trim={0 50 0 0},clip,width=0.48\textwidth]{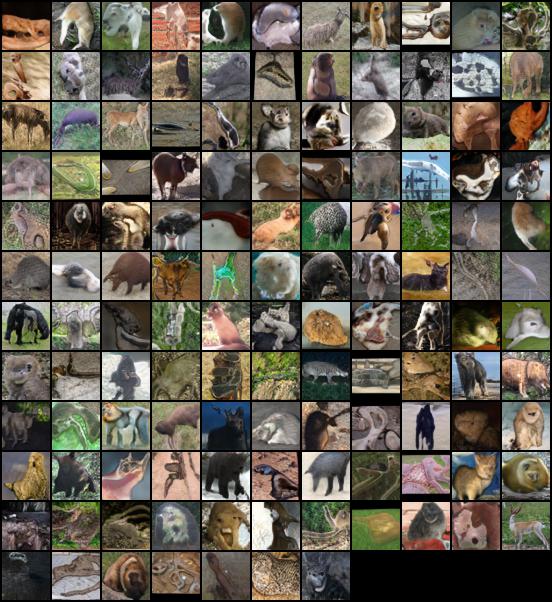}
\includegraphics[trim={0 50 0 0},clip,width=0.48\textwidth]{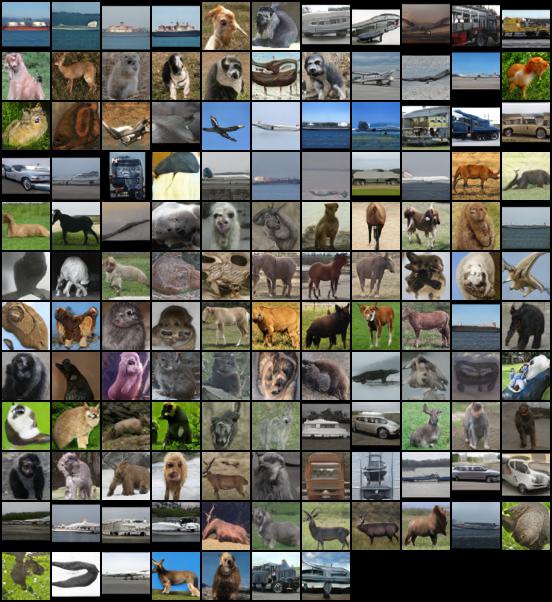}
\caption{Unsupervised image generation (without labels) on STL-10: (left) baseline WGAN; (right) GAEL. GAEL, which shares the same generator and discriminator like the baseline exhibits much more realistic images with meaningful semantic information. Images chosen randomly.}
\label{fig:STL10 samples}
\end{figure*}

Generative adversarial networks (GAN) are a powerful tool that allows learning data distributions and is widely used in many fields \cite{Dziugaite2015TrainingGN,karras2018progressive,hussein2019imageadaptive,SRGAN,Luc2016SemanticSU,Pix2Pix,nvidia_synthetic_GAN,Brophy2019QuickAE,Hartmann2018EEGGANGA,MaskGAN,JetchevBV16,LeakGAN,Brock2019LargeSG}. Despite their great success, GANs still suffer a few notable limitations.

First, the distribution of the generated data still does not match the distribution of the real one. In almost every application, GANs will generate at least some outputs that appear obviously synthetic. Many GAN training techniques are unstable, as they lack a mechanism that can autonomously stop the training upon a point of convergence, rather than upon some point decided by human intervention. The problem of mode collapse, or partial mode collapse, also still remains unsolved:  while GAN successfully generates some parts of the data distribution, there are other parts that it does not generate.  Furthermore, although the GAN generator can create high quality outputs, most GANs do not supply a mechanism that can take the output and infer its latent representation, which could be very helpful for unsupervised learning.
This last point constitutes a particularly acute limitation. If the network cannot infer a latent space, then it is excluded from a number of applications involving an encoder.

This paper explores whether it is possible to successfully incorporate an encoder with GAN. To do so, we present GMM-based generative adversarial encoder learning (GAEL), a hybrid GAN-Encoder network. While GAEL demonstrates improved performance on different GANs, its method is fairly simple. It consists of adding to the GAN an encoder, which shares most of its weights with the discriminator (or critic in WGAN). Then we train the encoder using a simple loss that is added to the regular GAN loss. The shared weights and simple loss mean that the network's innovations come with very little computational overhead or additional training time. After training finishes, we fit the outputted latent space of the real images to a Gaussian Mixture Model (GMM). Then we sample latent vectors from the GMM and use them as the generator input for generating synthetic images.

We evaluate GAEL on three important applications related to GANs: image reconstruction, image generation, and clustering. To examine how well our encoder learns meaningful representation, we decode the latent space images and compare the reconstructions to the originals (there is no use of GMM at this stage). Then, we use our encoder to learn a new distribution (GMM) to replace the original one used for the generator input during training. We then test how well the network generates images from the new latent distribution. Finally, we test how well our network clusters. We do this by fitting the GMM to the latent space, using GMM modes as proxies for dataset labels, i.e., each Gaussian in the GMM is assumed to represent a class in the data.

In our experiments, we use a variety of datasets including MNIST, CIFAR-10, CIFAR-100, CelebA, STL-10, and Imagenet. We use both quantitative and qualitative methods to evaluate the results. For metric assessment, we evaluate our image generation results with inception(IS) \cite{inception} and Frechet Inception Distance (FID) \cite{FID} scores, and our clustering results with the Adjusted Rand Index (ARI), Normalized Mutual Information (NMI), and clustering Accuracy (ACC) \cite{Xu15Comprehensive}.

On all datasets and evaluation metrics, our strategy performs very well. Our GMM-based GAN generates better images than those generated using its baselines and many other unsupervised image generation techniques. The network's clustering performance is competitive with the results of the CNN-based clustering and, as far as we know, better than current GAN-based techniques.

In summary, this paper introduces \our, a hybrid GAN-Encoder framework. The GAN and the Encoder train simultaneously, and the encoder and discriminator share weights throughout nearly the entire training. The method performs robustly on multiple datasets. We summarize GAEL’s contributions as follows:
(i) It improves how GAN performs at image reconstruction, clustering and image generation; (ii) the method is generic, and can be ported to other architectures;
(iii) the encoder could likely help improve future GAN implementations; and (iv) it stabilizes training.

\begin{figure*}[t]
\includegraphics[width=0.99\textwidth]{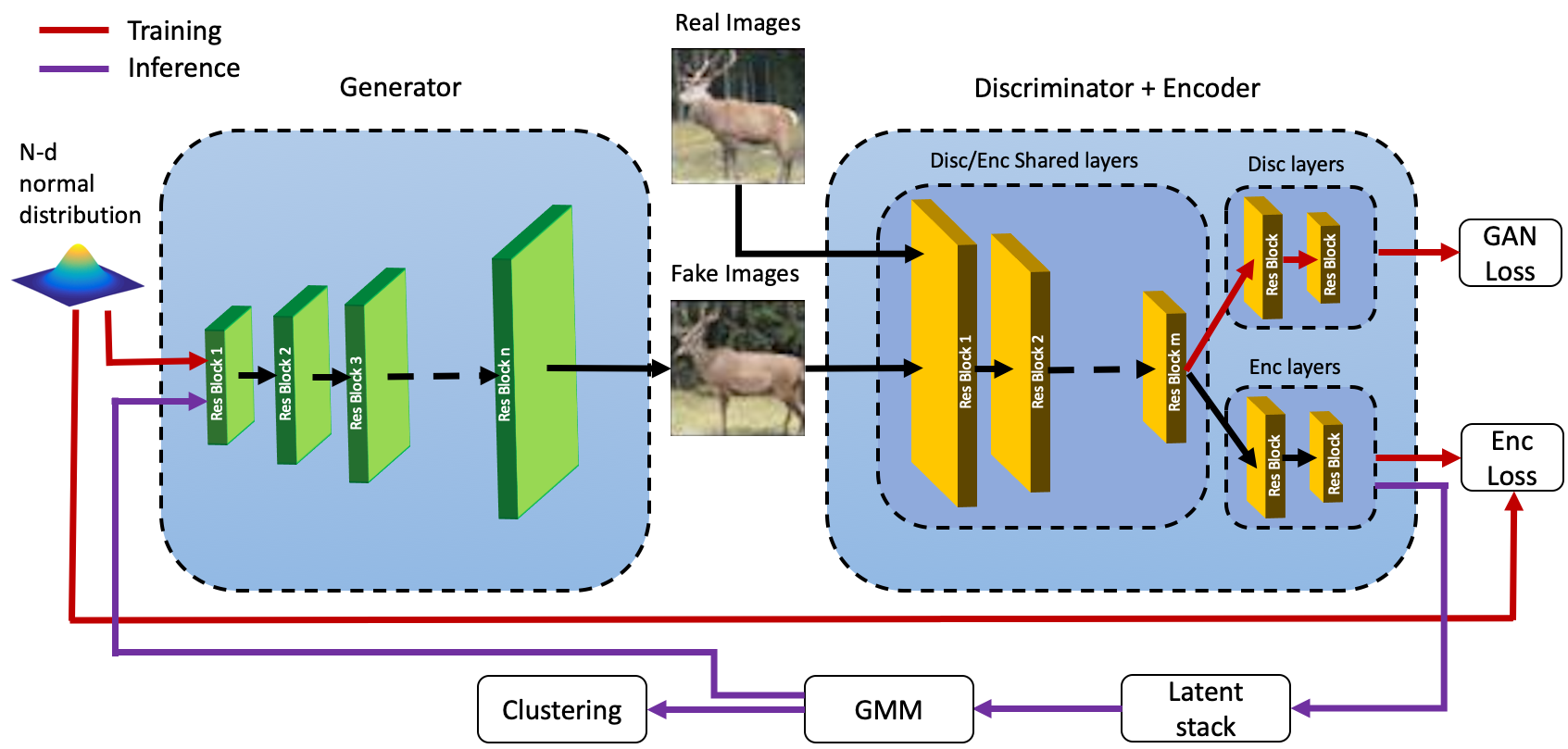}
\caption{\emph{Our proposed GAEL framework.} \our  combines the standard GAN architecture with an encoder. The encoder and discriminator train simultaneously, sharing  weights except for the last layers. There are two losses: regular GAN loss and the encoder's own loss. After the training finishes, we aggregate the latent space of the images and model the latent space with a GMM. Then the GMM distribution serves two purposes: 1) for clustering and 2) as an input distribution to the generator. }
\label{fig:framework}
\end{figure*}


\section{Related Work}


The research that has been performed on GANs is vast. Since the seminal work by Goodfellow et al \cite{GAN}, there have been significant attempts to improve the performance of GANs. The improvements have been developed through a variety of means, including through improved architectural structure \cite{BEGAN,BiGAN,LAPGAN,PROGAN,DCGAN,SAGAN,StyleGAN}, the introduction of labelled data to improve performance \cite{Mirza2014ConditionalGA,Kaneko2017GenerativeAC,Wang2018HighResolutionIS,Miyato2018cGANsWP,SSGAN}, and on to different loss functions \cite{Arjovsky2017WassersteinGA,GP-WGAN,Ni18CAGAN,CT-WGAN,Miyato2018SpectralNF,SphereGAN}. For example, some works add a maximum-mean-discrepancy loss to improve GAN training \cite{Liang17MMD, wang2019improving}. Other approaches use reconstruction loss to optimize the generator without a discriminator \cite{bojanowski2018optimizing,Hoshen18NAM}. Other approaches used architecture search to find the generator architecture \cite{Gong_2019_ICCV,doveh2019degas,Yuan20Off,wang2019agan}.

In general, GANs generate images from a given latent distribution. Yet, the other direction is usually not present, i.e., getting representation that leads to the generation of a given image as provided in auto-encoders \cite{VAE, donahue2019large}. Calculating such a representation is useful for unsupervised learning, e.g., using the latent codes of the images for clustering. 

While most GAN works do not provide such capability, several works have explored such directions. Since we propose such a modification to GANs, and show an improvement by doing so, we focus now on works that suggested similar modifications to GAN architectures and are the closest to our work. 

\begin{figure*}[t]
\centering
\includegraphics[width=0.32\textwidth]{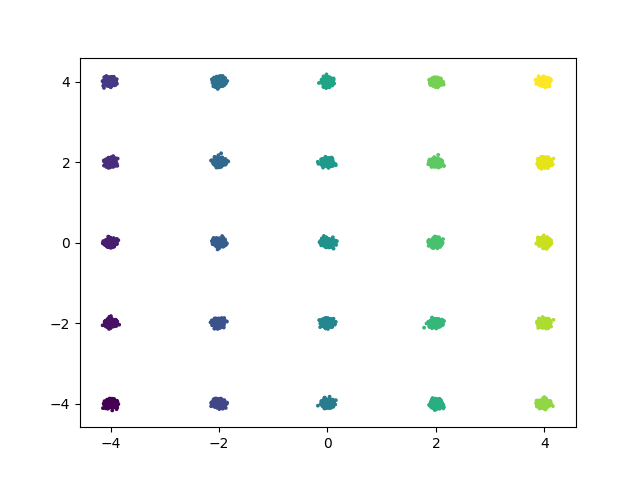}
\includegraphics[width=0.32\textwidth]{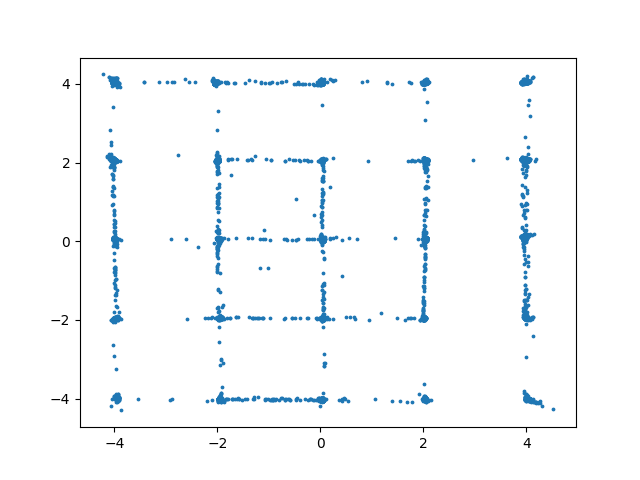}
\includegraphics[width=0.32\textwidth]{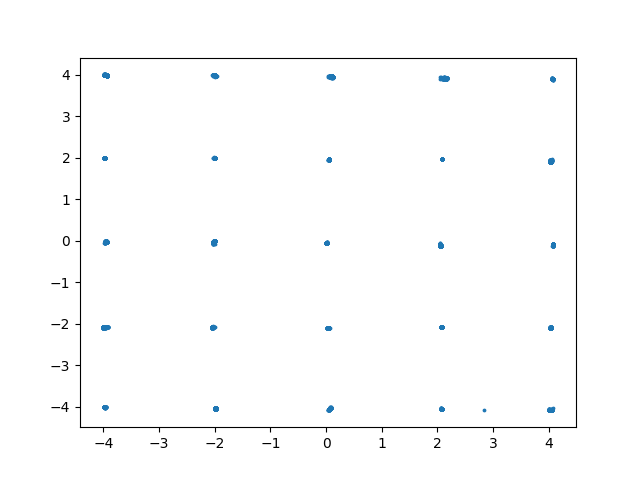}
\caption{From left to right: Original 5 X 5 toy dataset; WGAN-baseline generated data; and GAEL generated data. Note that while WGAN struggles to learn the data distribution, GAEL  uses the exact same architecture but with the addition of an encoder and GMM-modeling of the latent code generates the data much better. }
\label{fig:Toy_GAEL}
\end{figure*}

ALI \cite{ALI} suggested using another network for encoding and trained their framework such that it would jointly generate the images and the latent space. It learns to match the joint distribution of the real images with their encoded images, and the generated latent space along with the generated images. To improve the ALI approach, ALICE \cite{Chunyuan17ALICE} added a cycle consistent criteria, which used an encoder-generator mechanism to introduce a conditional entropy loss function between the real images and the reconstructed images. 

Notice that these works optimize the encoder in a way that enforces its output to have a normal distribution, which is less suited for clustering. In addition, trying to enforce a normal distribution of an output imposes a constraint upon the generator's ability to generate high quality images. The architecture is also more difficult to train, as the discriminator trains adversarially with not only the generator but also the encoder. This adds complexity to the already complex training mechanism of GANs. In our work, we propose an alternative path and do not enforce the distribution of the output of the encoder to be similar to the output of the latent distribution. Indeed, we require similarity between the two but in a weaker way. After that, we model the latent distribution with a GMM, which is very different than the actual (normal) distribution used in the training.

InfoGAN \cite{Chen2016InfoGANIR} maximized the mutual information between the latent space and the generated images. InfoGAN's results suggest that the network had learned interpretable latent representation. 
While InfoGAN enforces consistency on only a subspace of the latent space, our approach enforced consistency on the entire latent space. Learning the entire latent space reconstruction gave us the benefit of having an autoencoder. Furthermore, InfoGAN enforces the consistency between the input of the generator and the output of the discriminator through parametric optimization of both networks. Employing such simultaneous matching on both networks prevented the encoder from learning a good representation. This is because the flexibility of training both the generator and discriminator to minimize mutual information may lead to a trivial solution, e.g., encoding everything in one pixel.

VEEGAN \cite{VEEGAN} takes a different approach. It has two loss terms, one that tries to learn the inverse function of the generator back to the latent space, and a second that tries to minimize the cross entropy between the mapped real images and the prior latent space. 
VEEGAN enforces consistency across the entire latent space in the output of the encoder $(E(G(z)) - z)$, minimizing the network parameters of both the encoder and generator. Minimizing the encoder and generator together leads to the same issues that we described in InfoGAN. VEEGAN also enforces similarity between the original latent space and the output of the encoder's distribution, because it reduces the entropy between the original latent space and the output of the encoder. As with ALI, a loss function constrains the latent space, which hinders clustering and reduces the quality of the generated images. 

Our proposed framework relies also on using Gaussian Mixture Models (GMM) for the GAN latent space. 
Several previous approaches used mixtures with GANs. The work in \cite{Yang18Mixture} suggested using a GAN mixture for clustering. Another work compared GMM performance to the one of GANs for data modeling \cite{Richardson2018OnGA}. PacGAN \cite{Lin2018PacGANTP} provides the discriminator with multiple generated examples instead of one to improve the generated diversity and avoid mode collapse. A concurrent strategy \cite{farnia2020gatgmm} provides a novel GAN training that is based on the optimal transport theory for learning Gaussian mixture models. Another approach uses GMM for the input latent space of the GAN \cite{benyosef2018gaussian} and uses it in training. Our method, on the 
other hand, employs a standard Gaussian distribution during the training. Only at the end do we model the latent distribution as a GMM, using the entire dataset. We sample from the GMM only during inference.

\begin{table*}[t]\label{image-generation-results}
    \centering
    \begin{tabular}{|l|c|c|c|c|c|c|}
    \hline
    &\multicolumn{2}{|c|}{CIFAR10}&\multicolumn{2}{|c|}{CIFAR100}&\multicolumn{2}{|c|}{STL10 (48x48)}\\
    \hline
    &IS&FID&IS&FID&IS&FID\\
    \hline
    WGAN-GP \cite{GP-WGAN}&
    7.86&-&
    -&-&
    -&-\\
    \hline
    CT-GAN \cite{CT-WGAN}&
    8.12&-&
    -&-&
    -&-\\
    \hline
    SN-GAN \cite{Miyato2018SpectralNF}&
    8.22&21.7&
    -&-&
    9.1&40.1\\
    \hline
    AutoGAN \cite{Gong_2019_ICCV}&
    8.55&12.42&
    -&-&
    9.16&31.01\\
    \hline 
    progressive GAN \cite{Karras2018ProgressiveGO}&
    8.8&10.7&
    -&-&
    -&-\\
    \hline
    E$^2$GAN \cite{Yuan20Off}&
    8.51&11.26&
    -&-&
    9.51&25.35\\
    \hline
    
    Baseline WGAN&
    8.49&12.89&
    8.62&17.37&
    9.09&40.71\\
    \hline
    Baseline+Encoder WGAN(our)&
    8.76&8.76&
    8.86&13.42&
    10.00&27.44\\
    \hline
    GMM-10 WGAN(our)&
    9.15&{\bf 7.70}&
    9.26&13.98&
    10.39&23.39\\
    \hline
    GMM-50 WGAN(our)&
    {\bf 9.24}& 7.83&
    {\bf 9.46}&{\bf12.96}&
    {\bf10.75}&{\bf 22.57}\\
    \hline\hline
    BigGAN \cite{Brock2019LargeSG} (supervised)&
    9.22&14.73&
    -&-&
    -&-\\
    \hline
    MHingeGAN \cite{Kavalerov_2021_WACV} (supervised)&
    9.58&7.5&
    14.36&17.3&
    12.16&17.59\\
    \hline
    \end{tabular}
    \caption{GAN inception score and FID of our method compared to some baselines on three datasets; best results are highlight in \textbf{bold}. We present existing supervised GAN results only for reference. Notice that our GAN is unsupervised but achieves improved results compared to the BigGAN whose GAN architecture and loss are very similar to ours.}
    \label{tab:GAN_results}
\end{table*}

\section{The GAEL method}

Our method uses a hybrid-architecture, combining a GAN with an
encoder network. We call our method Generative Adversarial  Encoder Learning (GAEL). We outline our design in Figure~\ref{fig:framework}.
It uses a standard generator and discriminator that can be trained with any existing GAN technique. This makes our proposed strategy very generic. To the discriminator, we add an encoder head that shares the same first layers of the discriminator. The goal of the encoder head is to create a latent code for the discriminator’s input images. It is trained using only the fake images with a loss that requires the encoder’s code to match the latent code used in the generator. After the training finishes, we use a new latent distribution, namely, a mixture of Gaussians, as input to the generator. To learn this distribution, we adapt a GMM to all the latent codes generated by the encoder for the data. We also use this GMM to cluster the data.

Our proposed architecture has multiple benefits. The
encoder network enables us, naturally, to encode and decode images.
Additionally, because it transfers the image space to feature space,
the network provides a suitable output for clustering.
We also try to use the encoder network to improve the generator’s input distribution. By learning the real images’ latent space distribution, we can replace the normal distribution with the new learned distribution. 

To summarize, we test GAEL on three separate tasks: (i) autoencoding; (ii) generation; and (iii) clustering. We turn now to describe each part of our proposed framework and its utility in these tasks.

{\bf The encoder.} The generator $G$ maps each sample $Z \in R^M$ from the latent space to an image $X \in R^{NxN}$ in the image space, i.e., $x=G(z)$. While we can find the corresponding $x$ for each $z$ using $G$, the inverse is not always possible as $G$ is not necessarily invertible. To alleviate that, we train an encoder that learns the inverse function using probabilistic regression.
We choose the following likelihood for the inverse function:
\begin{eqnarray}
\label{maximum-liklihood}
P(z | x) \sim N\left(Enc\left(x\right), \Sigma \left(x\right)\right),
\end{eqnarray}
where $Enc\left(x\right)$ is the first encoder output, which estimates the latent vector that generated the input $x$, and $\Sigma(x)$ is another encoder output that serves as confidence or variance of the estimation of $Enc(x)$. The goal of the confidence output is twofold: (i) as mentioned above, having a strict condition in the training on the relationship between the generating latent code and the output of the encoder may harm training. The variance helps to mitigate this issue; (ii) this confidence might be useful in some tasks, e.g., using it as a weight in clustering (we do not use it in this work but this is a possibility for future work).

Our encoder shares most of its weights with the discriminator, with only the last few layers being independent. This helps the encoder learn meaningful features, while adding very little to the training time.
The encoder loss is the log likelihood of Equation~\eqref{maximum-liklihood}:
\begin{eqnarray}
&& \hspace{-0.5in} \log\left(P\left(z | x\right)\right) =  -0.5\log\left(\left( (2\pi)^M \left|\Sigma(x)\right|\right)\right) \\ &&
\nonumber
-0.5\left(z - Enc\left(x\right)\right)^T\Sigma^{-1} (x)\left(z - Enc\left(x\right)\right).
\end{eqnarray}

We also tried to enforce "consistency loss." We attempted this by passing the original images through an encoder--generator--encoder sequence, and asked for consistency between the first encoded images and the last ones. I.e., adding $min(E(G(E(X))) - E(X))$.
We tried variations of this set up: e.g. freezing the generator and training the encoder; freezing the encoder while training the generator; and freezing neither, and training both concurrently. All cases led to heavy degradation across all categories of network performance.
The failure of this consistency loss led us to the idea that the encoder’s training should interfere with GAN training as little as possible.
Our current setup allows this. Moreover, in our framework, the ability to use the encoder to reconstruct images does not compromise image generation quality but rather strengthens it.

{\bf GAEL loss function.} As the encoder is trained with the GAN as a whole, we combine encoder and GAN loss .
There are many varieties of the loss function for GAN. For the presentation simplicity, we explain our method on the loss function below. However, it can work on other loss functions as well (as we show in the experiments, where we use our loss also with the Wasserstein GAN).

\begin{eqnarray}
\label{eq:GAN loss}
\min_G\max_D V(D,G) &= & E_{x \sim P_{data}}\left[\log D\left(x\right)\right] \\ \nonumber &&+ E_{z \sim noise} \left[\log \left(1-D\left(G\left(z\right)\right)\right)\right].
\end{eqnarray}

This leads to the following loss function that is used for GAEL training:
\begin{eqnarray}
\label{eq:Encoder loss}
\min_G\max_{D,Enc,\sigma} V(D,G) + \lambda E_{x,z \sim P(x,z)}[\log\left(P\left(z | x\right)\right)],
\end{eqnarray}
where $\lambda$ is a hyperparameter that balances the weight of the encoder in the training. We set its value to $10$ in all experiments (the network performed similarly when we tried other values for $\lambda$ as well).

Clearly, we can plug our Encoder with its loss in any other GAN framework just by replacing $V(D,G)$ with the appropriate loss. For example, if we used the WGAN loss, then the encoder would share weights with the critic (WGAN’s alternative to the discriminator).

{\bf Autoencoder.} To use GAEL as an autoencoder, we simply calculate the encoding of the input image $x$ using $Enc(x)$. Then we reconstruct the image using the (trained) generator $G$, which serves as the decoder in this case.

{\bf Generation.} When we train the GAN, we use inputs that are drawn from a random Gaussian distribution. Thus, a straightforward way to generate new data with the generator is to draw random vectors from the same distribution and use them to generate new examples using $G$.

One of the drawbacks of GANs is that some of their outputs do not really belong to the true distribution. Therefore, we want to avoid these regions in the original latent distribution that leads to such outputs. To do so, we attempt to model the latent distribution of the original images by a parametric distribution. Specifically, we use GMM. To do that, we pass all the (true) data through the  encoder to get a feature space for each image. We then fit a GMM to the encoded features. Having this GMM, we can sample a latent vector from it, which is then used for input to the generator.

While other distributions could have been used, our focus in this work is on images. Images have a high degree of variability (e.g., different features and semantic content). To capture the multiple modalities of this data, we use a multi-modal distribution, the GMM.

{\bf Clustering.} GMM typically works well for clustering when the images within the clusters are in close Euclidean space (e.g., MNIST), but struggles to cluster images of highly varied semantic information \cite{Xu15Comprehensive}. To achieve better results while working with complex data, meaningful representation needs to be learned. Here is where deep learned features can be used \cite{Aljalbout18Clustering,clusteringSurvey}. In particular to our case, GAEL may use a GMM on the latent distribution generated by the encoder from the real data in order to cluster.

If the encoder learns meaningful representation, i.e., keeps similar semantic images together in the latent space, then the clustering can be done with traditional clustering algorithms (e.g., Kmeans and GMMs) on the latent space of the original images. As mentioned above, we use GMM. We calculate a GMM for the whole data and then assign each input image to the Gaussian that has its largest probability.

\begin{figure}
\centering
\includegraphics[width=0.48\textwidth]{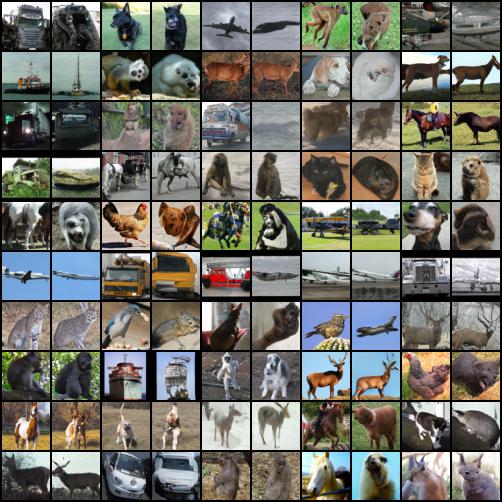}
\caption{Encoder-generator image reconstruction. Even columns contain the original images, odd contain reconstructions.}
\label{fig:Image reconstraction}
\end{figure}

{\bf Toy Example.}
We use a simple toy example to demonstrate our method. Figure \ref{fig:Toy_GAEL}(left) shows the original toy dataset, a GMM distribution with 25 components spread in a two-dimensional  grid pattern. On this toy dataset, we trained a baseline Wasserstein GAN and GAEL (i.e. the same GAN but with the addition of the encoder). Figure \ref{fig:Toy_GAEL}(middle) shows the results of the baseline generator. While it succeeded to recreate the original modes, it also left artifacts between them. These artifacts are hard to remove. We suggest this difficulty may owe to the fact that GAN architecture struggles to convert a smooth, i.e. normal, distribution to such a mode-disjoint one.

On the contrary, our method uses a GMM-distribution for our latent space, which is multi-modal and thus likely to be more adequate to the input latent distribution. Figure \ref{fig:Toy_GAEL}(right) exhibits our method. Note that in this case, the generated data matches the locations of the original data. More details on this experiment appear in the sup. material.

\begin{figure}
\centering
\includegraphics[width=0.5\textwidth]{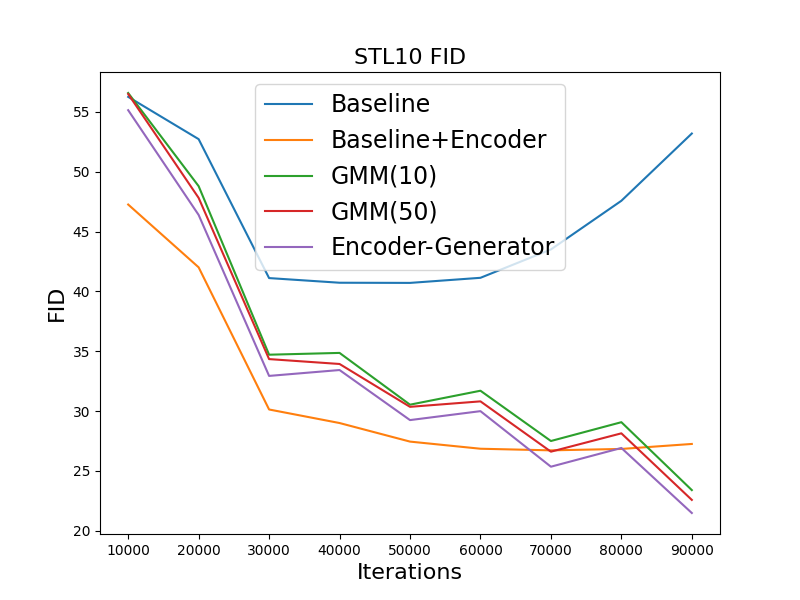}
\caption{FID score  across iterations trained on the STL-10 dataset. 
The baseline network performance degrades considerably as training progresses. 
Adding our encoder not only improves the scores, but also dramatically  reduces this performance degradation.  Incorporating the GMM further improves performance and stabilizes the training procedure. The encoder-generator network constitutes an upper bound for the GAEL performance. 
}
\label{fig:STL_convergence}
\end{figure}

\section{Experiments}
Except for STL-10, we used the BigGAN \cite{Brock2019LargeSG}  architecture for all datasets, with minimal modification. For the STL-10 dataset, we employ MHGAN \cite{Kavalerov_2021_WACV}. We also applied the same hyper-parameters throughout the experiments. We tested our method's performance on three different applications: image generation, image reconstruction, and clustering.

Our experiments use the following data sets to train and evaluate our method: 
CIFAR-10, CIFAR-100, STL-10, CelebA, MNIST and ImageNet. We present here quantitative results for the first three datasets on the clustering and image generation tasks. We also show some visual results in the paper, but more is covered in the supplementary material, which presents all the datasets. 

We model the $\Sigma(x)$ by diagonal matrix.
During our experiments, we found that learning the $\Sigma(x)$ lead to similar results as using $\Sigma(x)=I$, which made our method even simpler.

\subsection{Image reconstruction}
\label{experiment-reconstruction}

Our first experiment tests the visual quality of our encoder--generator reconstruction of the original images. This test was done by encoding the original images and reconstructing them with the generator. We then compare the originals with the reconstructions to evaluate whether our encoder had learned meaningful representation.
We present the original and reconstructed images side-by-side in Figure \ref{fig:Image reconstraction}. More examples on all datasets appear in the supplementary material.

The figures show that the network was able to reconstruct images with similar semantic information and with high quality. Note though that it does not reconstruct the same input image. This is a desirable property. We want to keep the same semantics, as we want the encodings to be useful for clustering. Yet, we do not want to have a strict tie between the input of the generator and the code of the encoder. We found found that adding such a strict constraint in training degrades the performance of the generated images significantly, as it limits the generator flexibility. This might be the reason for the lower image quality that was achieved in previous works that tried to have exact matches in the encoding. We are not aware of a previous work that is able to achieve the quality of reconstruction that we achieved here, with a generator that contains an encoder
(see the results on STL-10 and ImageNet, particularly challenging datasets).

\begin{table*}[t]
    \centering
    \begin{tabular}{|l|c|c|c|c|c|c|c|c|c|c|c|c|}
    \hline
    &\multicolumn{3}{|c|}{CIFAR10}&\multicolumn{3}{|c|}{CIFAR100}&\multicolumn{3}{|c|}{STL10}\\
    \hline
    &NMI&ARI&ACC&NMI&ARI&ACC&NMI&ARI&ACC\\
    \hline
    GAN \cite{Radford2016UnsupervisedRL}&0.265&0.176&0.315&0.120&0.045&0.151&0.2100&0.139&0.298\\
    \hline
    JULE \cite{JULE}&0.192&0.138&0.271&0.103&0.033&0.137&0.182&0.164&0.277\\
    \hline
    DEC \cite{DEC}&0.257&0.161&0.301&0.136&0.049&0.185&0.276&0.186&0.359\\
    \hline
    TUM \cite{TUMcluster}&-&-&0.325&-&-&-&-&-&\textbf{0.530}\\
    \hline
    DAC \cite{DAC}&0.396&0.306&0.522&0.185&0.088&0.238&0.366&0.257&0.470\\
    \hline
    {\small ClusterGAN}\cite{ClusterGAN}&$0.323$&$-$&$0.412$&$-$&$-$&$-$&$0.335$&$-$&$0.423$\\
    \hline
    Our&
    $\mathbf{0.443}$&$\mathbf{0.350}$&$\mathbf{0.555}$&
    $\mathbf{0.289}$&$\mathbf{0.135}$&$\mathbf{0.299}$&
    $0.377$&$0.230$&$0.383$\\
    \hline
    \end{tabular}
    \caption{Clustering results of various method on three datasets, best results highlight in \textbf{bold}}
    \label{tab:clustering results}
\end{table*}

\subsection{Image generation}\label{experiment-generation}
Our second experiment tested how well our network generated images. To measure generation quality, we used the inception \cite{inception} and FID \cite{FID} scores. For both metrics, we used the author's original code.

Table \ref{tab:GAN_results} presents scores for our baseline, essentially a modified version of the BigGAN code with the projection discriminator removed. We established our GAEL architecture by adding the encoder to this model. Our results demonstrate that we achieve significant improvement upon this baseline by adding our encoder and its loss function. 
This implies that our encoder is not only capable of encoding images (as demonstrated in Section~\ref{experiment-reconstruction}), but also improves GAN performance. We show the results with and without the GMM encoding of the latent space. 

Notice that we get improvement even before adding the GMM. We also show the impact of the number of Gaussians in the GMM on the performance. 

In the supplementary material, we show the generation results of GAEL based on vanilla GAN. Again, adding the encoder gives  a significant boost in performance, which is further improved with
the GMM-modeling of the latent space. This demonstrates that our method is generic and can be used with a variety of GANs and is not restricted to a specific GAN model. 

In Table \ref{tab:GAN_results}, we compare GAEL to other methods. We may divide these methods into two groups: the first, which does not use the labels included in the data set (unsupervised), and the second, which does use the data set labels(supervised). Clearly, having the labels provides a big advantage. Yet, using GAEL with GMM, we get better results than the BigGAN, which has our architecture, but uses labels. We also come very close to the results of the strong supervised technique, MHingeGAN \cite{Kavalerov_2021_WACV}. This shows the power of our approach, which does not use any labels during training but rather models the latent space via a GMM. Notice also our advantage over other unsupervised approaches. 

Note that BigGAN applies a "truncation trick" \cite{Brock2019LargeSG}, which truncates the normal distribution in order to achieve improved FID and Inception scores. This trick and the GMM modeling come from the same core idea---improving the latent space. However, while BigGan uses a trick to achieve this, including a truncation threshold hyper-parameter, our approach is more fundamental and based on direct learning from the data.

To establish an upper bound of GAEL’s ability to generate images, we measured the FID and inception scores on the reconstructions themselves. Because we generated these images from our encoded original images, we cannot of course compare these specific results with independent methods. However, these scores provided us an upper bound of what our encoder could achieve, and therefore provided a target that the network could aim at when generating images. We call this approach Encoder-Generator. 

We found that the encoder introduced stability to the training procedure. We measured the FID and IS scores of generated images across all iterations of the GAN, trained with and without the encoder. Figure~\ref{fig:STL_convergence} shows the FID results for STL-10. More examples appear in the supplementary material. 
While the scores of the network without the encoder degraded substantially as training progressed, GAEL experienced almost no loss at all. Note that the addition of the encoder greatly stabilized training  and that the results of GAEL with GMM are not far from the results of the upper bound Encoder-Generator. 





\subsection{Clustering}

Our final experiment tested our method's ability to cluster. To evaluate our clustering, we fit the GMM to the latent space, this time with the number of mixture models being equal to the number of labels in each dataset. Then each image was assigned to one of the mixture models. 
We evaluated the performance of our clustering algorithm using the Adjusted Rand Index (ARI), Normalized Mutual Information (NMI), and clustering Accuracy (ACC).
A qualitative demonstration of our clustering can be found in the supplementary materials.

Table \ref{tab:clustering results} shows that our results exceed the performance obtained by all other GAN-based clustering. On the CIFAR-10 and CIFAR-100 benchmarks, GAEL's performance, as measured by all metrics, delivers the best results compared to all methods. On the STL-10 dataset, two CNN-based (not GAN-based) approaches, TUM and DAC, outperform our clustering results.

\section{Conclusion}
This paper presents \our, a simple and effective approach to enhancing GAN performance by incorporating an encoder into its architecture. \our's simplicity derives from its low number of variables and simple loss function, which has only one hyperparameter. GAEL works robustly for a wide range of values of this parameter. Additionally, our encoder can be incorporated into almost any GAN architecture with negligible computational or training time cost. Beyond its performance improvements, it also adds a secondary benefit of stabilizing the training procedure.

Our results exhibit the encoder's utility in three separate applications: image reconstruction, generation, and clustering. We tested the architecture on a variety of different datasets using both metrical and qualitative analysis. In all cases, \our achieved high performance. We also tried to continue training GAEL with the new latent space of the GMM. This showed a small improvement, which was not significant enough to be presented here. Yet, a more sophisticated training could have benefited from this combination and we defer this to a future study. Another interesting direction is adapting our technique to the tasks of semi-supervised and few-shot learning. We hope that our proposed methodology will be found beneficial for many other applications and for improving other approaches to GAN.

\paragraph{Acknowledgement} We would like to thank Matt Dodson, for his writing advice and editorial oversight.



{\small
\bibliographystyle{ieee_fullname}
\bibliography{egbib}
}
\end{document}


\title{GMM-Based Generative Adversarial Encoder Learning \\ Supplementary Material}

\author{%
  Yuri Feigin, Hedva Spitzer and Raja Giryes \\
  School of Electrical Engineering\\
  Tel Aviv University\\
  Israel, 69978 \\
  \texttt{{yurifeigin@mail,raja@tauex}.tau.ac.il} \\
  }

\maketitle

\begin{abstract}
Here we present some additional examples for the main paper.
\end{abstract}
\section{Toy example}

\begin{figure*}[t]
\centering
\begin{subfigure}{.4\textwidth}
    \centering
    \includegraphics[width=0.95\textwidth]{Images/gmm_original_grid.png}
    \caption{}
\label{fig:toy grid}
\end{subfigure}
\begin{subfigure}{.4\textwidth}
    \centering
    \includegraphics[width=0.95\textwidth]{Images/gmm_samples.png}
    \caption{}
\label{fig:toy gen}
\end{subfigure}
\begin{subfigure}{.4\textwidth}
    \centering
    \includegraphics[width=0.95\textwidth]{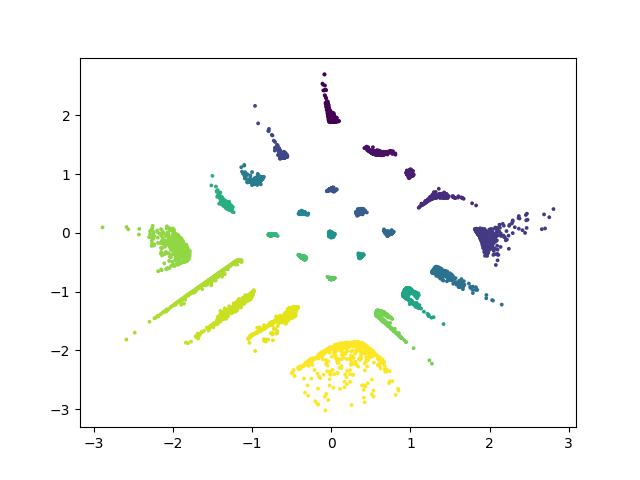}
    \caption{}
\label{fig:toy grid enc}
\end{subfigure}%
\begin{subfigure}{.4\textwidth}
    \centering
    \includegraphics[width=0.95\textwidth]{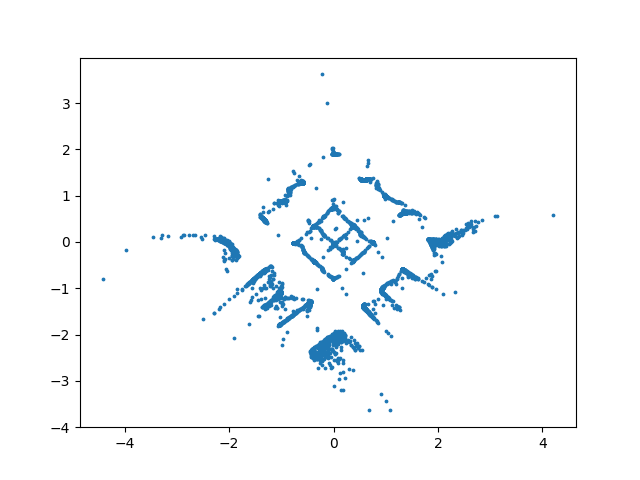}
    \caption{}
\label{fig:toy gen enc}
\end{subfigure}
\caption{Top left shows the original 5 x 5 toy dataset. Bottom left, the encoded original dataset. Top right, the WGAN baseline generated data. Bottom right, the encoded generated data.}
\label{fig:toy enc}
\end{figure*}

A toy example may provide the reader some intuition about how our method functions. As this demonstration is on paper, here we use a two-dimensional latent space in scatter grid. We projected both the original samples(figure \ref{fig:toy grid}) and the generated images (from a normal distribution) (figure \ref{fig:toy gen}) into two-dimensional space(figures \ref{fig:toy grid enc} and \ref{fig:toy gen enc}, respectively). Comparing the two latent spaces, we find that, while the main modes are represented fairly clearly, there is also evidence of object mixing in the latent space, as  indicated by the artefacts between the modes. Modelling the latent space of the original samples with the GMM reduces these artefacts, as we discussed in the paper.

The reader may also note that this new latent space is not distributed normally, because there are many regions between the modes without any density.

\section{Image reconstruction}
\begin{figure*}[t]
\centering
\includegraphics[width=0.90\textwidth]{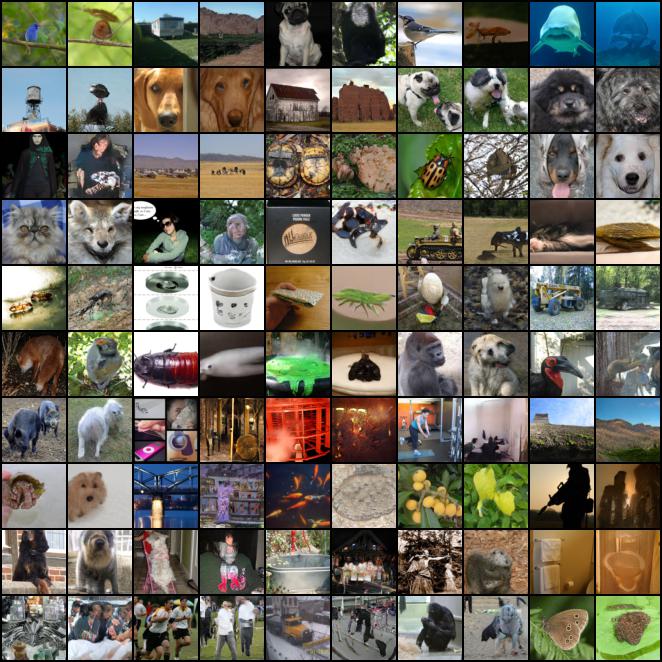}
\caption{Encoder-generator Imagenet 64x64 image reconstruction.  Even columns contain the original images, odd contain reconstructions. Random sample.}
\label{fig:I64_image_reconstruction}
\end{figure*}

\begin{figure*}[t]
\centering
\includegraphics[width=0.90\textwidth]{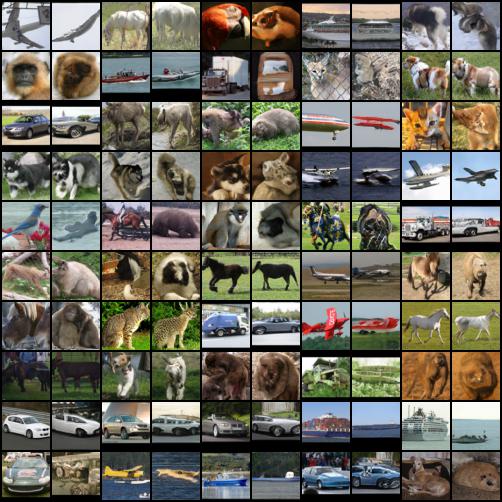}
\caption{Encoder-generator STL10 48x48 image reconstruction.  Even columns contain the original images, odd contain reconstructions. Random sample.}
\label{fig:STL10_image_reconstruction}
\end{figure*}

\begin{figure*}[t]
\centering
\includegraphics[width=0.90\textwidth]{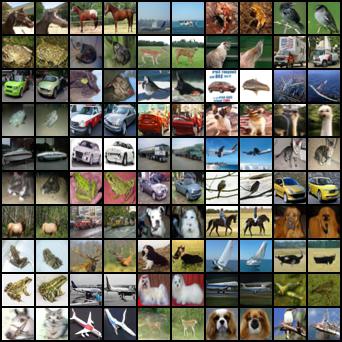}
\caption{Encoder-generator CIFAR10 image reconstruction.  Even columns contain the original images, odd contain reconstructions. Random sample.}
\label{fig:CIFAR10_image_reconstruction}
\end{figure*}

\begin{figure*}[t]
\centering
\includegraphics[width=0.90\textwidth]{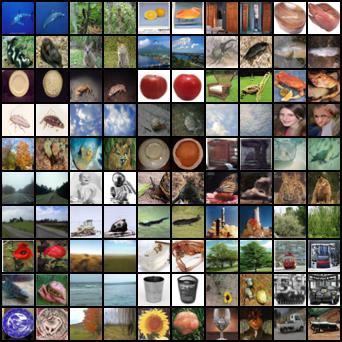}
\caption{Encoder-generator CIFAR100 image reconstruction.  Even columns contain the original images, odd contain reconstructions. Random sample.}
\label{fig:CIFAR100_image_reconstruction}
\end{figure*}

\begin{figure*}[t]
\centering
\includegraphics[width=0.90\textwidth]{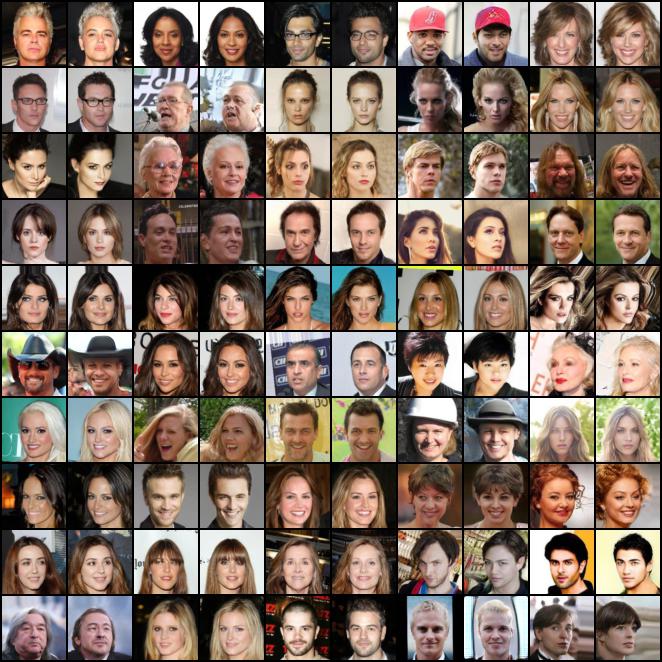}
\caption{Encoder-generator CelebA image reconstruction.  Even columns contain the original images, odd contain reconstructions. Random sample.}
\label{fig:CelebA_image_reconstruction}
\end{figure*}

\begin{figure*}[t]
\centering
\includegraphics[width=0.90\textwidth]{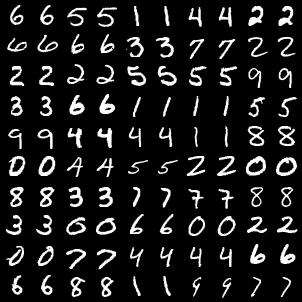}
\caption{Encoder-generator MNIST image reconstruction.  Even columns contain the original images, odd contain reconstructions. Random sample.}
\label{fig:MNIST_image_reconstruction}
\end{figure*}

In figures \ref{fig:I64_image_reconstruction}, \ref{fig:STL10_image_reconstruction}, \ref{fig:CIFAR10_image_reconstruction}, \ref{fig:CIFAR100_image_reconstruction}, \ref{fig:CelebA_image_reconstruction}, \ref{fig:MNIST_image_reconstruction}, we show more results of image reconstruction on the datasets MNIST, CIFAR10, CIFAR100, STL10, CelebA, and Imagenet.
We used relative small batch size of 256 images to train the Imagenet.

For all datasets, the images reconstructed well, retaining the originals’ semantic information. However the Imagenet results are more mixed. Although the network seemed to handle certain objects well (e.g. dogs, buildings, and cars), it struggled to handle other classes. Despite admitted imperfections, we think the results are promising.

While our attempts with ALI struggled with MNIST, GAEL seems to have handled the dataset well.

\section{Image generation}
This section presents GAEL’s image generation results compared to two baselines, WGAN and Vanilla GAN.
\subsection{WGAN (as in main paper)}
\begin{figure*}[t]
\centering
\includegraphics[width=0.90\textwidth]{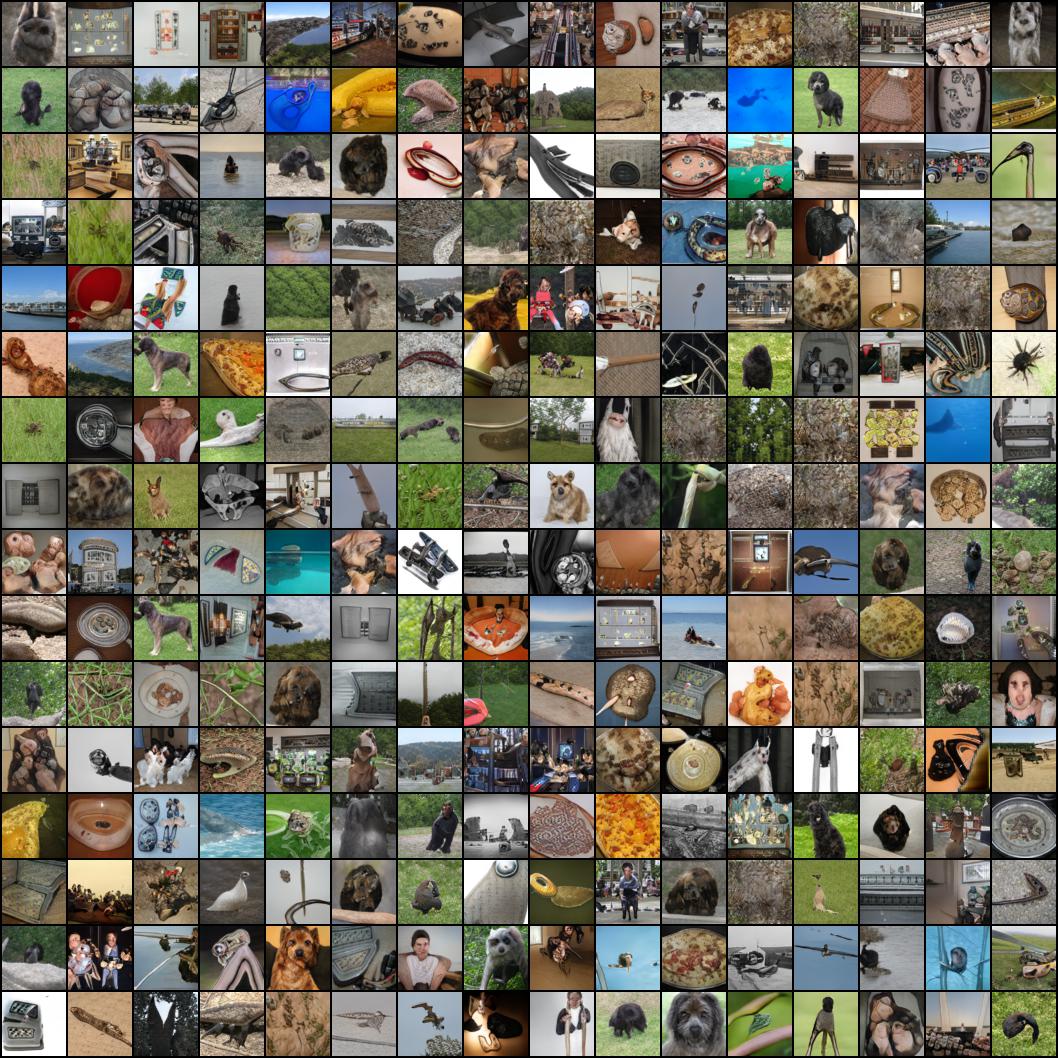}
\caption{Imagenet 64x64 image generation with baseline. Randomly sampled. Compare to Figure~\ref{fig:I64_image_generation_GAEL}.}
\label{fig:I64_image_generation_baseline}
\end{figure*}

\begin{figure*}[t]
\centering
\includegraphics[width=0.90\textwidth]{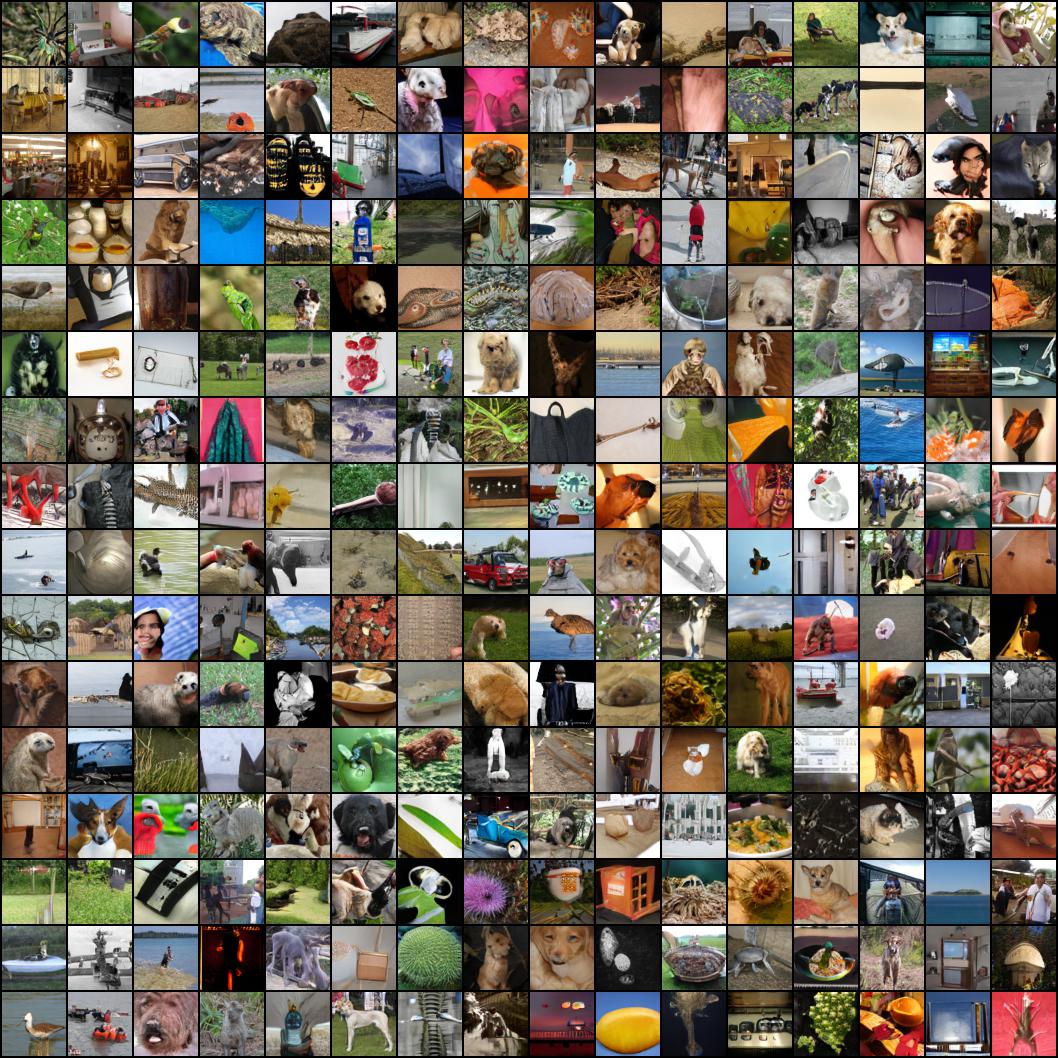}
\caption{Imagenet 64x64 image generation with GAEL. Randomly sampled. Compare to Figure~\ref{fig:I64_image_generation_baseline}.}
\label{fig:I64_image_generation_GAEL}
\end{figure*}

\begin{figure*}[t]
\centering
\includegraphics[trim={0 34 0 0},clip,width=0.90\textwidth]{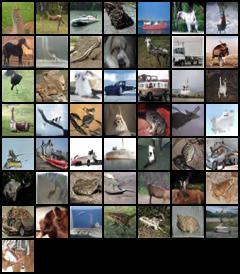}
\caption{CIFAR10 image generation with baseline. Randomly sampled. Compare to Figure~\ref{fig:CIFAR10_image_generation_GAEL}.}
\label{fig:CIFAR10_image_generation_baseline}
\end{figure*}

\begin{figure*}[t]
\centering
\includegraphics[trim={0 34 0 0},clip, width=0.90\textwidth]{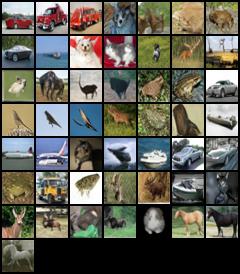}
\caption{CIFAR10 image generation with GAEL. Randomly sampled. Compare to Figure~\ref{fig:CIFAR10_image_generation_baseline}.}
\label{fig:CIFAR10_image_generation_GAEL}
\end{figure*}

\begin{figure*}[t]
\centering
\includegraphics[trim={0 34 0 0},clip,width=0.90\textwidth]{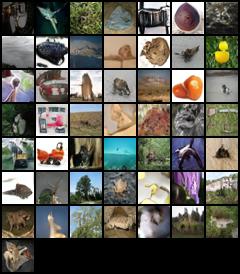}
\caption{CIFAR100 image generation with baseline. Randomly sampled. Compare to Figure~\ref{fig:CIFAR100_image_generation_GAEL}.}
\label{fig:CIFAR100_image_generation_baseline}
\end{figure*}

\begin{figure*}[t]
\centering
\includegraphics[trim={0 34 0 0},clip, width=0.90\textwidth]{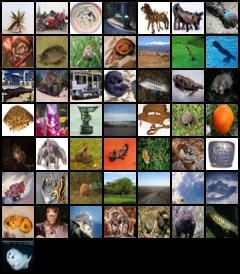}
\caption{CIFAR100 image generation with GAEL. Randomly sampled. Compare to Figure~\ref{fig:CIFAR100_image_generation_baseline}.}
\label{fig:CIFAR100_image_generation_GAEL}
\end{figure*}

\begin{figure*}[t]
\centering
\includegraphics[trim={0 66 0 0},clip, width=0.90\textwidth]{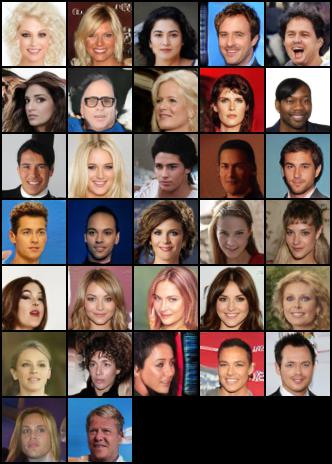}
\caption{CelebA image generation with GAEL. Randomly sampled.}
\label{fig:CelebA_image_generation_GAEL}
\end{figure*}

\begin{figure*}[t]
\centering
\includegraphics[trim={0 30 0 0},clip, width=0.90\textwidth]{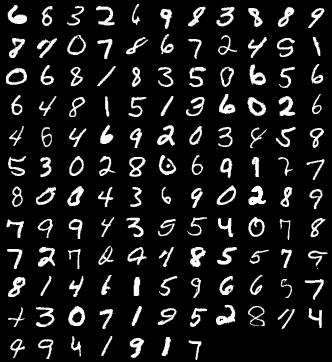}
\caption{MNIST image generation with GAEL. Randomly sampled.}
\label{fig:MNIST_image_generation_GAEL}
\end{figure*}

In figures \ref{fig:I64_image_generation_baseline}, \ref{fig:I64_image_generation_GAEL}, \ref{fig:STL10_image_generation_baseline}, \ref{fig:STL10_image_generation_GAEL}, \ref{fig:CIFAR10_image_generation_baseline}, \ref{fig:CIFAR10_image_generation_GAEL}, \ref{fig:CIFAR100_image_generation_baseline}, \ref{fig:CIFAR100_image_generation_GAEL}, \ref{fig:CelebA_image_generation_GAEL}, \ref{fig:MNIST_image_generation_GAEL}, we present the image generation performances of the baseline WGAN and GAEL. We invite the reader to compare their results on each dataset. Because we trained our Imagenet Dataset on a resolution of 64 x 64, we do not have a reference to compare the results there to any other methods. Nonetheless we wish to summarize the results here.  The baseline IS and FID scores were 13.14 and 33.03, respectively. For the baseline with encoder loss, the results were 14.60 and  20.42. Lastly, the results of the GMM were 14.78 and 22.66. We can see that GAEL improved the results over the baseline.

\subsection{Vanilla GAN}
Table \ref{tab:Vanilla_GAN_results} compares our results with a baseline, \emph{Vanilla GAN} whose architecture was based on BigGAN and which used the original GAN loss. The other rows show our method’s progressive modifications upon Vanilla GAN. First, by adding an encoder, then by fitting GMMs of ten and fifty. On each data set, our method consistently improved the results of the baseline. While on the CIFAR10 and CIFAR100 datasets, the GMM-50/Encoder combination yielded the best results, for the STL-10 data set,  the encoder GAN (without the GMM) fared better. A possible explanation lies in the varied nature of the STL-10 dataset. If the Vanilla GAN struggled to generate images on it, then the encoder would not work well either.

\begin{figure*}[t]
\centering
\includegraphics[trim={0 50 0 0},clip,width=0.90\textwidth]{Images/stl10_baseline.jpg}
\caption{STL10 48x48 image generation with baseline. Randomly sampled. Compare to Figure~\ref{fig:STL10_image_generation_GAEL}.}
\label{fig:STL10_image_generation_baseline}
\end{figure*}

\begin{figure*}[t]
\centering
\includegraphics[trim={0 50 0 0},clip,width=0.90\textwidth]{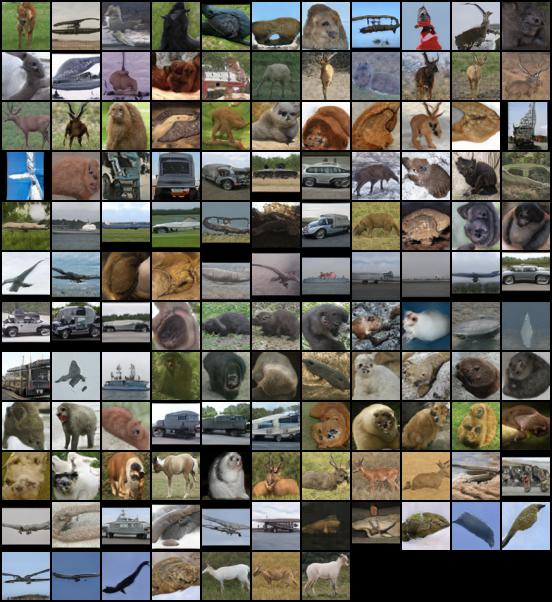}
\caption{STL10 48x48 image generation with GAEL. Randomly sampled. Compare to Figure~\ref{fig:STL10_image_generation_baseline}.}
\label{fig:STL10_image_generation_GAEL}
\end{figure*}

\begin{table*}[b]
    \centering
    \begin{tabular}{|l|c|c|c|c|c|c|}
    \hline
    &\multicolumn{2}{|c|}{CIFAR10}&\multicolumn{2}{|c|}{CIFAR100}&\multicolumn{2}{|c|}{STL10 (48x48)}\\
    \hline
    &IS&FID&IS&FID&IS&FID\\
    \hline
    Baseline Vanilla GAN&
    7.68&34.23&
    7.75&39.1&
    6.57&66.93\\
    \hline
    Baseline+Encoder Vanilla GAN(our)&
    7.33&28.51&
    6.97&33.19&
    {\bf8.04}&{\bf 43.64}\\
    \hline
    GMM-10 Vanilla GAN(our)&
    8.20&20.64&
    7.98&28.26&
    7.48&55.80\\
    \hline
    GMM-50 Vanilla GAN(our)&
    {\bf 8.24}& {\bf 20.16}&
    {\bf 8.00}&{\bf27.98}&
    7.72&54.39\\
    \hline
    \end{tabular}
    \caption{Vanilla GAN inception score and FID of our method compared to some baselines on three datasets; best results are highlighted in \textbf{bold}.}
    \label{tab:Vanilla_GAN_results}
\end{table*}